\documentclass[10pt,twocolumn,letterpaper]{article}

\usepackage{iccv}              % To produce the CAMERA-READY version
\definecolor{iccvblue}{rgb}{0.21,0.49,0.74}
\usepackage[pagebackref,breaklinks,colorlinks,allcolors=iccvblue]{hyperref}

\def\paperID{*****} % *** Enter the Paper ID here
\def\confName{ICCV}
\def\confYear{2025}

\title{TRACE: Textual Reasoning for Affordance Coordinate Extraction}

\author{
Sangyun Park\textsuperscript{1,*} \qquad
Jin Kim\textsuperscript{2,3,}\thanks{Equal contribution.} \qquad
Yuchen Cui\textsuperscript{4,5} \qquad
Matthew S. Brown\textsuperscript{2,3,}\thanks{Corresponding author.}
\\[+0.8em]
\textsuperscript{1}ABB Robotics, Seoul, Republic of Korea \\
\textsuperscript{2}Department of Radiology, University of California, Los Angeles, CA, USA \\
\textsuperscript{3}Center for Computer Vision and Imaging Biomarkers, University of California, Los Angeles, CA, USA \\
\textsuperscript{4}Computer Science Department, University of California, Los Angeles, CA, USA \\
\textsuperscript{5}Robot Intelligence Lab, University of California, Los Angeles, CA, USA \\
\vspace{1em}
{\tt\small \{sang-yun.park@kr.abb.com, kimjin116@g.ucla.edu, yuchencui@cs.ucla.edu, mbrown@mednet.ucla.edu\}}
}

\begin{document}
\maketitle
\begin{abstract}
Vision-Language Models (VLMs) struggle to translate high-level instructions into the precise spatial affordances required for robotic manipulation. While visual Chain-of-Thought (CoT) methods exist, they are often computationally intensive. In this work, we introduce \textbf{TRACE (Textual Reasoning for Affordance Coordinate Extraction)}, a novel \textbf{methodology} that integrates a textual \textbf{Chain of Reasoning (CoR)} into the affordance prediction process. We use this methodology to create the \textbf{TRACE dataset}, a large-scale collection created via an autonomous pipeline that pairs instructions with explicit textual rationales. By fine-tuning a VLM on this data, our model learns to externalize its spatial reasoning before acting. Our experiments show that our \textbf{TRACE-tuned model} achieves state-of-the-art performance, reaching \textbf{48.1\%} accuracy on the primary Where2Place (W2P) benchmark (a 9.6\% relative improvement) and \textbf{55.0\%} on the more challenging W2P(h) subset. Crucially, an ablation study demonstrates that performance scales directly with the amount of reasoning data used, confirming the CoR's effectiveness. Furthermore, analysis of the model's attention maps reveals an interpretable reasoning process where focus shifts dynamically across reasoning steps. This work shows that training VLMs to generate a textual CoR is an effective and robust strategy for enhancing the precision, reliability, and interpretability of VLM-based robot control.
Our dataset and code are available at \url{https://github.com/jink-ucla/TRACE}.
\end{abstract}   
\section{Introduction}
\label{sec:intro}

The quest to build generalist robots capable of performing diverse manipulation tasks has been significantly advanced by the advent of Vision-Language Models (VLMs) and Vision-Language-Action (VLA) models~\cite{team2503gemini, zhao2025cot}. These models show immense promise by leveraging vast commonsense knowledge to interpret natural language instructions and visual scenes for high-level task planning~\cite{huang2023voxposer, liang2022code}. However, a critical gap remains between high-level reasoning and the precise, low-level spatial understanding required for physical manipulation~\cite{pan2025omnimanip, zhao2025manipbench}. Effectively grounding ambiguous language commands in the physical world of robot affordances is a fundamental challenge~\cite{ahn2022can, song2025robospatial}, limiting the reliability of VLM-driven robots in unstructured environments.

Recent research has explored several avenues to bridge this gap. One popular approach is to have VLMs predict intermediate action representations, such as affordance keypoints, which serve as targets for low-level controllers~\cite{yuan2024robopoint, huang2024rekep, liu2024moka, xu2025a0}. While effective, these methods often treat the VLM as a black-box predictor, without an explicit, interpretable reasoning process. Concurrently, a new wave of research has focused on enabling VLMs to ``think visually" through a visual Chain-of-Thought (CoT). These approaches allow models to generate intermediate visual states, such as future goal images~\cite{zhao2025cot, li2025imagine}, edit the input image to refocus attention~\cite{fu2025refocus, shao2024visual}, or perform guided visual search to find relevant details~\cite{wu2024v, chung2025don, xu2025language, zhang2025mllms}. These methods demonstrate that iterative, multi-step visual processing is crucial for complex reasoning. However, they can be computationally intensive and often rely on external tools or generative models, creating a complex pipeline for robotic control.

\begin{figure*}[t]
    \centering
    \includegraphics[width=1\textwidth]{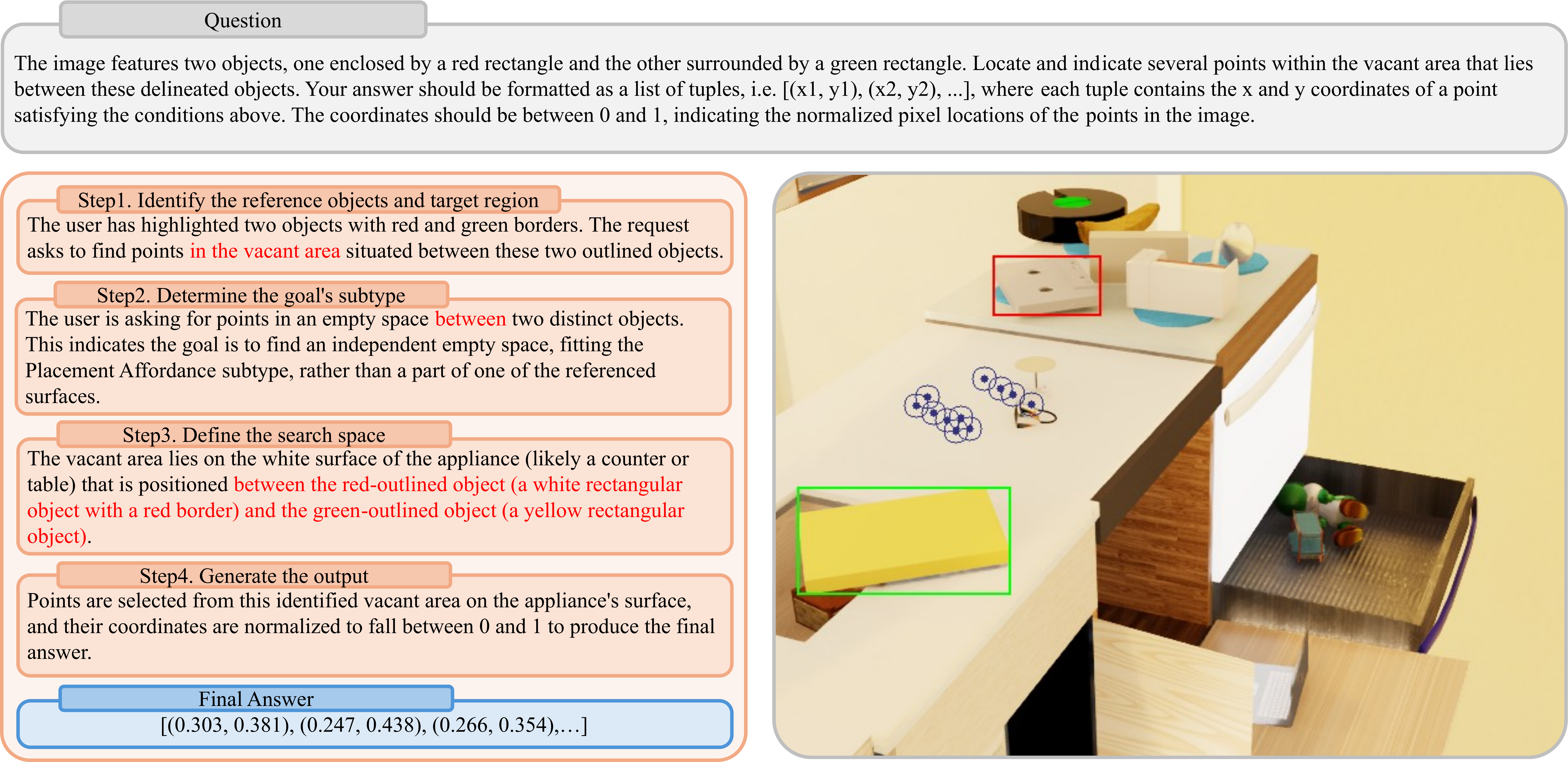}
    \caption{An example data point from our TRACE reasoning dataset illustrating its overall structure. Each entry consists of an image and a corresponding natural language question that requires spatial reasoning, such as finding the vacant area between the two delineated objects. The dataset also provides the explicit, multi-step reasoning process required to solve the instruction. This process includes identifying reference objects, determining the goal's subtype as a ``Placement Affordance", defining the search space on the appliance's surface, and generating the final coordinates as a list of tuples.}
    \label{fig:placement_affordance}
\end{figure*}

In this work, we argue that for a VLM to generate precise and reliable low-level actions, it must not only predict what to do but also reason about why it is doing it. Inspired by the success of CoT prompting in enhancing the reasoning capabilities of LLMs~\cite{wei2022chain}, and its recent application in robotics~\cite{hao2025cora, hao2025embodied}, we introduce a framework that integrates a textual `Chain of Reasoning' (CoR) into the spatial affordance prediction process, as illustrated in \Cref{fig:placement_affordance}. Our approach differs from visual CoT methods by focusing on generating an interpretable textual rationale for a grounded visual action. This strategy directly leverages the VLM's linguistic strengths and is computationally simpler than generating intermediate images.

\begin{figure*}[t]
    \centering
    \includegraphics[width=1\textwidth]{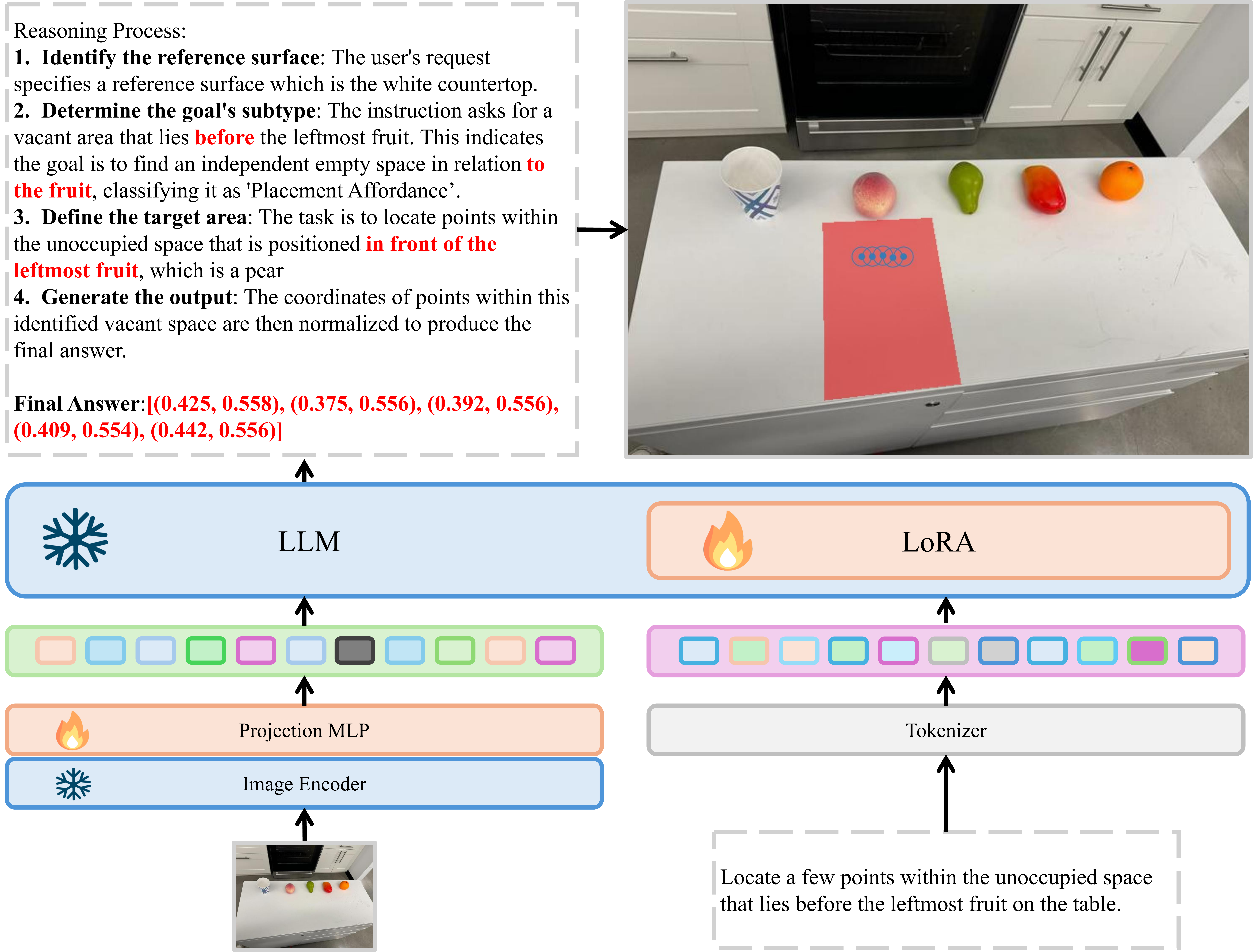}
    \caption{Overview of the model's reasoning pipeline. Given an image and a corresponding natural language instruction, the system begins a multi-step reasoning process. The model first determines the subtype of the goal and establishes the relevant reference surface based on the image. It then defines the target area by interpreting the request. This process utilizes an Image Encoder, Tokenizer, and a Large Language Model (LLM). Finally, a Projection MLP generates the output, which consists of normalized coordinates for points within the identified vacant space.}
    \label{fig:reasoning_process}
\end{figure*}

To achieve this, we introduce the \textbf{TRACE (Textual Reasoning for Affordance Coordinate Extraction) methodology}. This methodology includes a new data generation pipeline used to create the large-scale \textbf{TRACE dataset}, created by enhancing the RoboPoint~\cite{yuan2024robopoint} data generation pipeline. Our pipeline now automatically generates explicit textual reasoning steps that justify the selection of ground-truth action points. By fine-tuning a VLM on this augmented dataset, we teach it to externalize its reasoning process, leading to more accurate and robust predictions for low-level manipulation.

\begin{itemize}
    \item{We introduce the \textbf{TRACE methodology} and the resulting large-scale \textbf{TRACE dataset}, which pairs language instructions with both keypoint affordances and a CoR, created via a modified autonomous data generation pipeline. }
    \item{We demonstrate that instruction-tuning a VLM with our CoR data significantly improves its spatial understanding and the precision of its predicted action points for robotic manipulation tasks.}
    \item{Our experiments show that our \textbf{TRACE-tuned model (RoboPoint+TRACE)} achieves \textbf{48.1\%} accuracy on the challenging Where2Place benchmark, a statistically significant \textbf{9.6\% relative improvement} over the original RoboPoint model, showcasing a more interpretable and reliable path for VLM-based control. }
\end{itemize}

\section{Related Work}
\label{sec:rw}

Our work is situated at the intersection of VLM-driven robotic manipulation, spatial affordance prediction, and CoT reasoning. We build upon methods for learning visual affordances while drawing a sharp contrast with the emerging field of visual CoT.

\subsection{Chain-of-Thought for Robotics and Vision}
The concept of CoT prompting, introduced by Wei \etal~\cite{wei2022chain}, demonstrated that prompting large language models to generate intermediate textual steps significantly improves their performance on complex reasoning tasks. This paradigm has been foundational, inspiring a new direction in how models approach problem-solving. Our work adapts this core idea, which we term a textual CoR, to the visuomotor domain of spatial grounding for robotics.

Recently, similar textual reasoning strategies have been applied to robotics. For instance, Hao \etal have explored a ``Chain of Robotic Actions Reasoning" to generate sequences of high-level actions for manipulation tasks~\cite{hao2025cora} and humanoid locomotion~\cite{hao2025embodied}. These methods show that breaking down a complex instruction into a textual plan improves task success. Our work shares this emphasis on textual, step-by-step reasoning. However, instead of generating high-level action plans, we apply the reasoning process to a different problem: justifying the selection of precise, low-level spatial coordinates for manipulation. Our CoR provides an explicit rationale for \textit{why} a specific point in space is chosen, directly linking high-level reasoning to low-level physical grounding.

\subsection{Contrasting with Visual CoT}
In parallel, a significant body of work has explored a \textbf{Visual CoT}, where reasoning is externalized through intermediate visual states rather than text. A prominent approach is the generation of future images that act as subgoals. For example, CoT-VLA~\cite{zhao2025cot} generates images of anticipated future scenes, while other work focuses on visualizing intermediate steps to guide the reasoning process~\cite{li2025imagine}.

Other forms of Visual CoT involve editing or interacting with the input image. REFOCUS~\cite{fu2025refocus} generates code to programmatically edit the visual input, such as by masking or highlighting regions, to focus the model's attention. Similarly, other methods enable the model to perform interactive reasoning by selectively revisiting parts of an image~\cite{chung2025don} or conducting a guided visual search to find missing details~\cite{wu2024v}.

While these visual and interactive methods are powerful, they often introduce significant computational overhead from image generation, editing, or multi-step visual processing. Our textual CoR framework is proposed as a lightweight yet effective alternative. By grounding the reasoning process in language—a native modality for VLMs—we improve the precision of spatial predictions without the complexity and computational cost associated with generating or manipulating visual data. Our approach leverages the model's innate linguistic capabilities to directly enhance its spatial understanding.

\begin{figure*}[ht]
    \centering
    \includegraphics[width=1\textwidth]{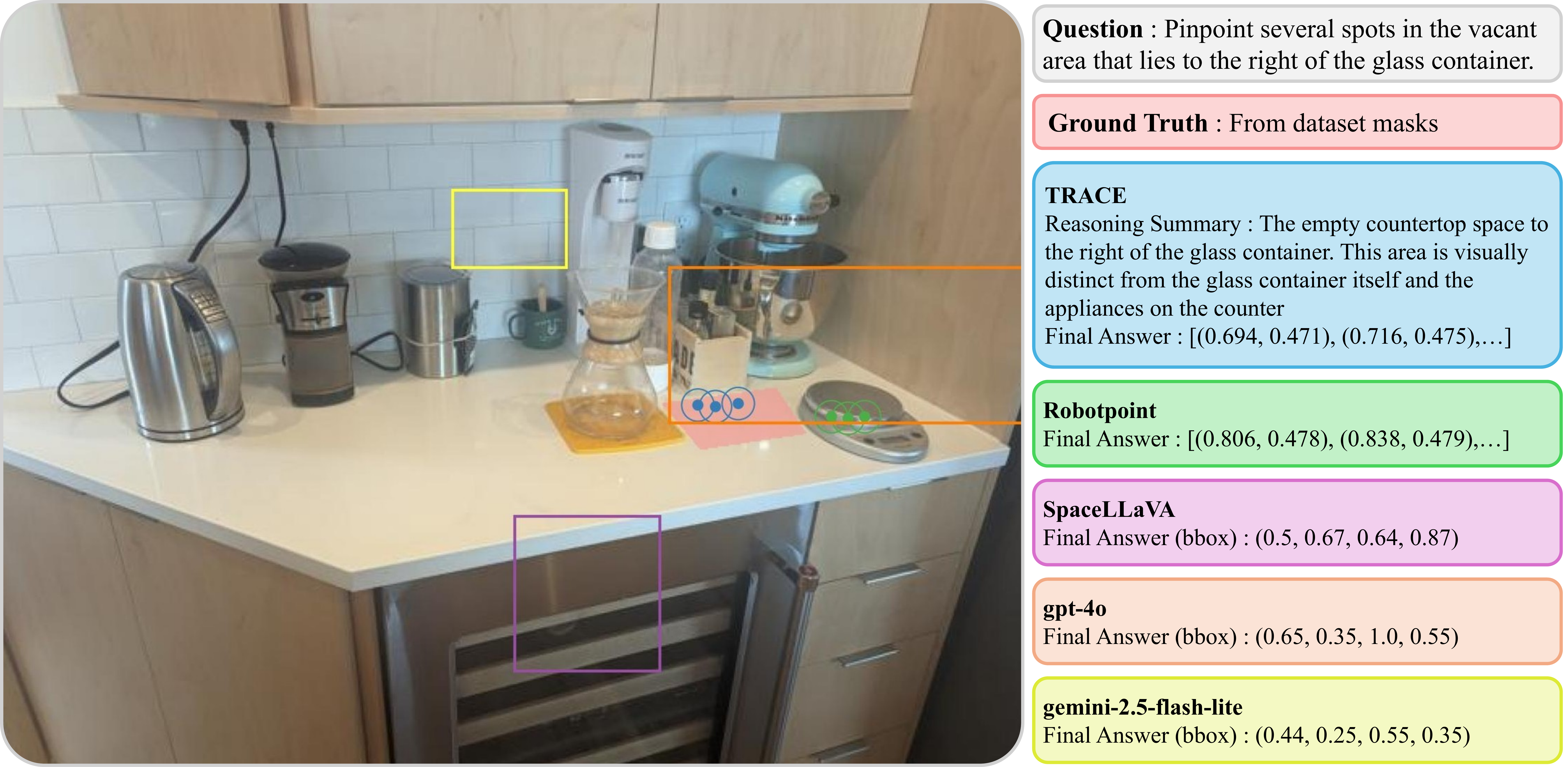}
    \caption{A qualitative comparison of TRACE with other leading models on a sample from our reasoning dataset. The given instruction is to ``Pinpoint several spots in the vacant 
area that lies to the right of the glass container".}
    \label{fig:qualitative_comparison}
\end{figure*}

\section{Method}
\label{sec:method}

Our approach enhances a VLM's ability to ground language in the physical world by fine-tuning it on our novel \textbf{TRACE} dataset. This dataset is specifically designed to integrate a textual \textbf{CoR} into the learning process, teaching the model not only to predict precise spatial affordances but also to articulate the reasoning behind them. This section details the model architecture, the dataset construction, and the optimization strategies employed.

\subsection{Model Architecture}
Our model architecture is composed of three core components: a base language model, a vision encoder, and a multimodal projector. For our primary results, we use \textbf{Vicuna-v1.5-13B}~\cite{chiang2023vicuna} as the large language model. \textbf{For computationally intensive experiments, such as our ablation study and attention visualization, we utilize a smaller \textbf{Vicuna-v1.5-7B} variant to facilitate analysis.} The vision backbone is a pretrained \textbf{CLIP-ViT-Large-Patch14-336}~\cite{radford2021learning} model, where we extract visual features from the penultimate layer to capture rich semantic information. These visual tokens are then mapped into the language model's embedding space via a \textbf{2-layer MLP projector} that uses a GELU activation function. To ensure memory-efficient and fast computation, the model implementation leverages \textbf{Flash Attention 2}~\cite{dao2023flashattention}. All prompts and responses adhere to the standard \texttt{Vicuna v1} conversation template.

\begin{table*}[t]
\centering
\caption{Main results on spatial affordance prediction benchmarks. The metric is the percentage of predicted points falling within the ground-truth mask, averaged over three runs. Our model, TRACE, outperforms all baselines across all benchmarks.}
\label{tab:main_results}
\begin{tabular}{lccc}
\toprule
\textbf{Model} & \textbf{RoboRefIt} & \textbf{W2P} & \textbf{W2P (h)} \\
\midrule
RoboPoint(FFT)+\textbf{TRACE} & \textbf{42.9\% $\pm$ 0.8} & \textbf{48.1\% $\pm$ 0.1} & \textbf{55.0\% $\pm$ 3.5} \\
RoboPoint(FFT) & 41.7\% $\pm$ 0.6 & 43.9\% $\pm$ 0.6 & 46.9\% $\pm$ 4.2 \\

\midrule
RoboPoint(LoRA)+\textbf{TRACE} & \textbf{48.1\% $\pm$ 2.8} & \textbf{43.7\% $\pm$ 4.1} & \textbf{41.2\% $\pm$ 7.3} \\
RoboPoint(LoRA) & 40.6\% $\pm$ 3.0 & 36.1\% $\pm$ 1.3 & 30.7\% $\pm$ 0.2 \\

\midrule
SpaceLLaVA~\cite{lu2023vl} & 20.0\% $\pm$ 0.5 & 15.0\% $\pm$ 1.6 & 13.6\% $\pm$ 2.1 \\
GPT-4o~\cite{hurst2024gpt} & 6.5\% $\pm$ 0.8 & 18.7\% $\pm$ 2.6 & 17.8\% $\pm$ 4.8 \\
Gemini & 5.2\% $\pm$ 0.1 & 7.8\% $\pm$ 0.2 & 6.6\% $\pm$ 0.2 \\
\bottomrule
\end{tabular}
\end{table*}

\subsection{TRACE Dataset and Preprocessing}

The cornerstone of our work is the \textbf{TRACE dataset}, a large-scale collection of \textbf{200,000 training samples} designed to teach VLMs explicit reasoning. The dataset is a composite of two data sources: 100,000 novel reasoning-augmented samples and 100,000 standard visual instruction-tuning samples to match the data scale of the original RoboPoint model~\cite{yuan2024robopoint}.

The first component, which is our primary contribution, consists of \textbf{100,000 samples with an explicit textual Chain of Reasoning (CoR)}. To create these, we took the image and instruction pairs from the original RoboPoint datasets and used the Gemini API to synthetically generate a step-by-step rationale for each one. This process, detailed in Appendix \ref{sec:dataset_details}, prompted the model to deconstruct the task into distinct cognitive steps: identifying reference objects, determining the goal's subtype (e.g., ``Placement Affordance"), defining the target area, and justifying the final coordinate selection, as illustrated in \Cref{fig:placement_affordance}.

The second component consists of an additional \textbf{100,000 samples} from the LVIS~\cite{gupta2019lvis} and VQA~\cite{liu2024improved} datasets. These samples follow a standard instruction-following format without the explicit CoR and ensure that our model is trained on a dataset of comparable size to the baseline, isolating the effect of our reasoning-augmented data.

The learning task for the entire dataset is formulated as an autoregressive prediction problem, where the model must generate both a textual rationale (for the TRACE-specific data) or a direct answer, followed by target coordinates, as shown in \Cref{fig:reasoning_process}. Given an input image and a language instruction, the model is trained to generate a response that first outputs the text and then provides the target 2D coordinates, normalized as $\{(x_i, y_i) \mid x_i, y_i \in [0, 1]\}$. This approach teaches the model to connect its reasoning directly to its final spatial prediction.

To manage this large-scale dataset efficiently, we implement several key preprocessing and data-loading optimizations. We enable \textbf{lazy preprocessing} to process data on the fly, significantly reducing memory consumption. All images are \textbf{padded to a square aspect ratio} to create uniform input for the vision encoder. To further improve training efficiency, we optimize batching by \textbf{grouping samples by modality length}, which minimizes the amount of padding required. Finally, to prevent data-loading bottlenecks, the pipeline is supported by \textbf{12 dataloader workers}.

\subsection{Training and Optimization}
We employ two distinct training strategies corresponding to our two model variants.

\textbf{For our main 13B model}, whose results are reported in \Cref{tab:main_results}, we perform \textbf{full fine-tuning (FFT)} for one epoch on the \textbf{TRACE dataset}. All weights of the language model are updated during this process to achieve maximum performance.

\textbf{For the 7B model}, used for the ablation study and attention analysis presented in \Cref{fig:ablation_study} and \Cref{fig:reasoning_attention}, we use a parameter-efficient approach. Specifically, we employ \textbf{Low-Rank Adaptation (LoRA)}~\cite{hu2022lora} with a rank of $r=128$ and a scaling factor of $\alpha=256$. This allows us to adapt the model by updating only the LoRA adapters applied to the language model's linear layers.

Both training configurations use the \textbf{AdamW optimizer}~\cite{loshchilov2017decoupled} with a global learning rate of $2 \times 10^{-6}$ and a \textbf{cosine annealing scheduler} with a 3\% warmup period. To ensure memory efficiency, we utilize \textbf{bfloat16 (bf16)} mixed-precision and enable \textbf{gradient checkpointing}.

\section{Experimental Results}
\label{sec:results}

We evaluate our model on challenging spatial affordance prediction tasks to demonstrate that integrating a textual CoR significantly improves performance. Our experiments are designed to answer: 1) How does our \textbf{TRACE-tuned model} perform against state-of-the-art VLMs and the original RoboPoint? 2) Is the performance gain statistically significant?

\subsection{Benchmarks}
\textbf{Benchmarks:} We evaluate performance on two real-world benchmarks:
\begin{itemize}
    \item \textbf{RoboRefIt:} A 250-image dataset featuring cluttered scenes where objects can only be distinguished by relational language instructions~\cite{lu2023vl}.
    \item \textbf{Where2Place (W2P):} A challenging 100-image dataset for identifying free space based on relational language. It includes a difficult subset, \textbf{Where2Place (h)}, with 30 examples containing relation types not seen during training~\cite{yuan2024robopoint}.
\end{itemize}
\textbf{Baselines:} We compare our model against several strong baselines:
\begin{itemize}
    \item \textbf{RoboPoint:} The original VLM from Yuan et al.~\cite{yuan2024robopoint}, which serves as our primary baseline.
    \item \textbf{GPT-4o:} A state-of-the-art proprietary VLM used in a zero-shot prompting setup~\cite{hurst2024gpt}.
    \item \textbf{SpaceLLaVA:} An open-source VLM specialized for spatial reasoning tasks~\cite{liu2023visual}.
    \item \textbf{Gemini:} A state-of-the-art proprietary VLM used in a zero-shot prompting setup~\cite{team2503gemini}.
\end{itemize}

\begin{figure*}[t]
    \centering
    \includegraphics[width=1\textwidth]{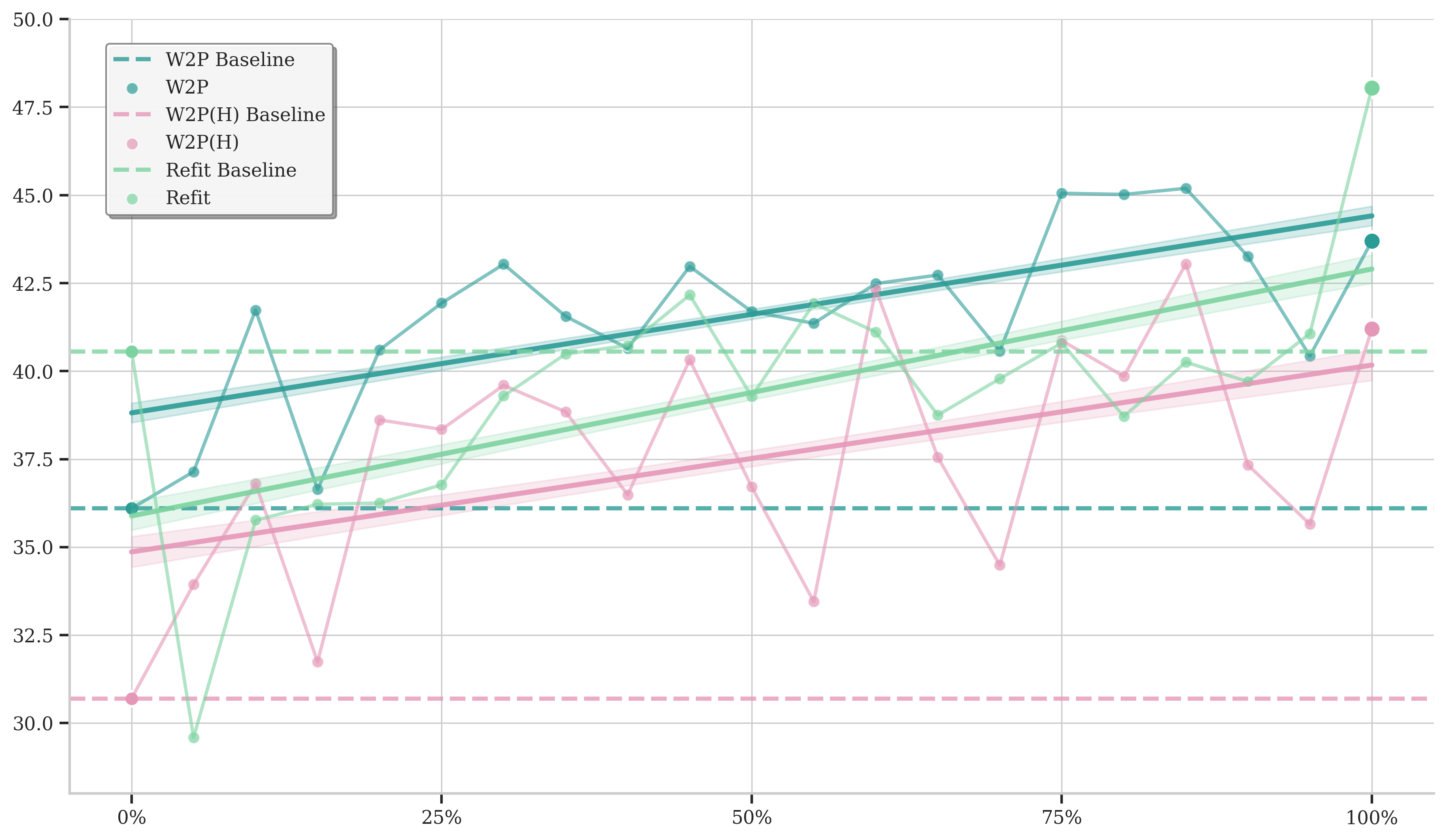}
    \caption{Ablation study on the impact of reasoning data. The plot shows the performance on the RoboRefIt, Where2Place (W2P), and Where2Place (h) benchmarks as the percentage of the TRACE reasoning dataset used for training is increased from 0\% (baseline) to 100\%. The solid lines represent the trend (linear regression), while the shaded areas indicate the 95\% confidence interval. Performance across all tasks consistently improves with more reasoning data, providing strong evidence for the effectiveness of our approach.}
    \label{fig:ablation_study}
\end{figure*}

\begin{figure*}[tbp]
    \centering
    \includegraphics[width=1\textwidth, height=17cm]{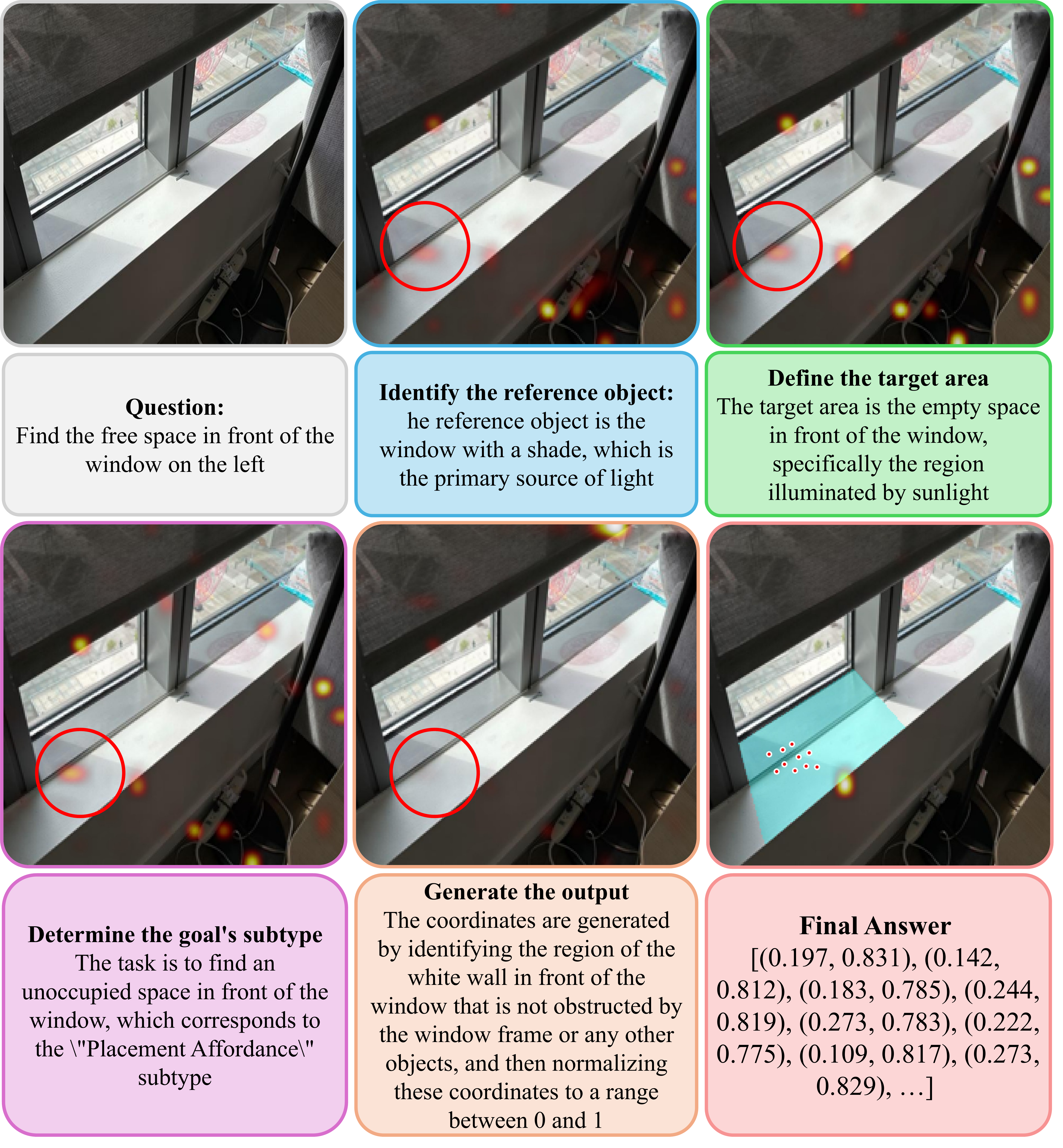}
    \caption{A visualization of the model's reasoning attention map for the instruction: ``Find the free space in front of the window on the left." The figure illustrates the model's focus across the four-step textual reasoning process. The attention heatmap is overlaid on the input image. (1) \textbf{Identify Reference Object} and (2) \textbf{Define Target Area}: The model exhibits diffuse, weak attention during the initial steps of identifying the window and defining the target area. (3) \textbf{Determine Goal's Subtype}: Attention begins to \textbf{focus} as the model classifies the task. A \textbf{distinct high-attention region (bright orange) emerges} over the target area, suggesting this is a critical reasoning step. (4) \textbf{Generate Output}: In the final step, there is minimal attention on the image, especially over the predicted points (shown as dots), indicating the model relies on its completed textual reasoning to generate the final coordinates rather than direct, concurrent visual evidence.}
    \label{fig:reasoning_attention}
\end{figure*}

\subsection{Performance on Spatial Affordance Prediction}
Our primary finding is that instruction-tuning with the \textbf{TRACE dataset} leads to consistent improvements in spatial affordance prediction across all benchmarks. As shown in Table~\ref{tab:main_results}, our \textbf{TRACE-tuned model (RoboPoint+TRACE)} achieves \textbf{48.1\% accuracy on the Where2Place benchmark}, a notable improvement over the 43.9\% accuracy of the original RoboPoint. This represents a relative improvement of 9.6\%.

To provide a qualitative understanding of these performance differences, Figure~\ref{fig:qualitative_comparison} illustrates a direct comparison on a challenging sample from our reasoning dataset, where our \textbf{TRACE-tuned model} correctly identifies the target area while other models struggle.

The performance gains are even more pronounced on the more challenging \textbf{Where2Place (h)} subset, where our \textbf{TRACE-tuned model} outperforms the RoboPoint by a significant margin (55.0\% vs. 46.9\%). These results demonstrate that the explicit textual reasoning process helps the model better resolve the ambiguity in complex, unseen relational instructions. Both our \textbf{TRACE-tuned model} and the original RoboPoint significantly outperform other specialist and generalist VLMs, highlighting the strength of the underlying architecture and data generation pipeline.

\subsection{Ablation Study on Reasoning Data}
To conclusively demonstrate that our primary contribution—the textual Chain of Reasoning (CoR) data—is responsible for the performance improvements, we conducted an ablation study. For this analysis, we utilized the 7B LoRA variant of our model to efficiently assess the impact of data scaling. We trained the model on progressively larger subsets of the TRACE dataset, from 0\% (the baseline model trained without any CoR data) to 100\% (the full 200,000-sample dataset).

The results, presented in \Cref{fig:ablation_study}, confirm a strong, positive correlation between the quantity of reasoning data and model performance. Specifically, as the training data scales from 0\% to 100\%, performance on \textbf{RoboRefIt} improves by \textbf{7.5} points (from 40.6\% to 48.1\%), on \textbf{Where2Place (W2P)} by \textbf{7.6} points (from 36.1\% to 43.7\%), and on \textbf{Where2Place (h)} by a substantial \textbf{10.5} points (from 30.7\% to 41.2\%). The positive linear trend for each benchmark is statistically significant, confirming a robust dose-dependent relationship between CoR data and performance.

Notably, the performance on the challenging \textbf{Where2Place (h)} subset, which contains unseen relational concepts, exhibits the steepest improvement, with a \textbf{34.2\% relative gain} over its baseline. This result strongly indicates that exposure to the explicit reasoning process is particularly crucial for enhancing the model's ability to generalize to novel and more complex instructions. The consistent upward trend across all benchmarks validates that the CoR is not just an explanatory artifact but a vital component for improving spatial grounding.

\subsection{Analysis of the Reasoning Mechanism}
\label{sec:analysis}

To better understand \textit{why} fine-tuning with the TRACE dataset improves performance, we analyzed the model's internal state by visualizing its attention map during the reasoning process. This provides insight into how the model utilizes the textual Chain of Reasoning (CoR) to solve spatial tasks.

\Cref{fig:reasoning_attention} presents such a visualization for the instruction,``Find the free space in front of the window on the left." The attention heatmap reveals a dynamic process. During the initial steps—(1) ``Identify Reference Object" and (2) ``Define Target Area"—the model's attention is weak and diffuse as it establishes a general context. However, attention becomes more \textbf{focused} during step (3), ``Determine Goal's Subtype." While still diffuse, a \textbf{new, distinct high-attention region (the bright orange spot) emerges} over the general target area under the window. This suggests that classifying the instruction is a critical juncture where the model solidifies its understanding of the task.

Crucially, in the final step (4), ``Generate Output," there is almost no visual attention focused on the area where the final coordinates are predicted. This observation supports our central hypothesis: the model leverages the completed textual CoR as its primary guide for action generation, rather than relying on continuous, intensive visual grounding. The reasoning chain acts as a powerful intermediate representation that bridges the gap between ambiguous language and precise spatial coordinates, directly contributing to the improved accuracy shown in \Cref{tab:main_results}.

\subsection{Statistical Validation}
To validate the significance of our improvements over the original RoboPoint, we performed a two-sample t-test. The analysis shows that the performance gain on the primary \textbf{Where2Place (W2P) benchmark is statistically significant ($p=0.022 < 0.05$)}. This confirms that the CoR provides a meaningful improvement for identifying affordances in free space. While the improvements on the RoboRefIt ($p=0.36$) and the challenging Where2Place (h) ($p=0.27$) subsets were not statistically significant at the $p < 0.05$ level, our model consistently outperformed the baseline across all categories, demonstrating the general effectiveness of our approach.

\section{Conclusion}
\label{sec:conclusion}

In this work, we introduced TRACE, a framework that enhances spatial affordance prediction by integrating an explicit textual Chain of Reasoning (CoR). We demonstrated through comprehensive experiments that fine-tuning a VLM on our large-scale TRACE dataset enables it to generate more accurate and interpretable low-level actions. Our approach's success is supported by three key findings: 1) Our model achieves state-of-the-art performance, outperforming strong baselines with a statistically significant \textbf{9.6\% relative improvement} on the primary Where2Place benchmark. 2) Our ablation study proves that these gains are directly attributable to the CoR data, as performance scales with the quantity of reasoning examples provided during training. 3) Our analysis of attention maps provides qualitative evidence of an interpretable, multi-step reasoning process, validating our method's design. Together, these results show that training for textual reasoning is a lightweight yet powerful method for improving the grounding of language in robotic actions.

\subsection{Limitations and Future Work}
While promising, our work has limitations. The reasoning chains in our dataset are synthetically generated and may not capture the full complexity of human thought processes. Additionally, while our attention analysis provides insight, the model lacks a mechanism to explicitly control this process or report confidence estimates for its predictions.

For future work, we plan to explore more sophisticated and less-structured reasoning generation. We also aim to extend the CoR framework to a wider range of robotic tasks, including multi-step manipulation and navigation. Finally, investigating methods to leverage the insights from attention maps to further improve model reliability represents a promising avenue for creating more capable and trustworthy VLM-driven robots.
{
    \small
    \bibliographystyle{ieeenat_fullname}
    \bibliography{main}
}

\clearpage
\setcounter{page}{1}
\maketitlesupplementary

\section{Additional Qualitative Results}
\label{sec:qualitative}

\begin{figure*}[ht]
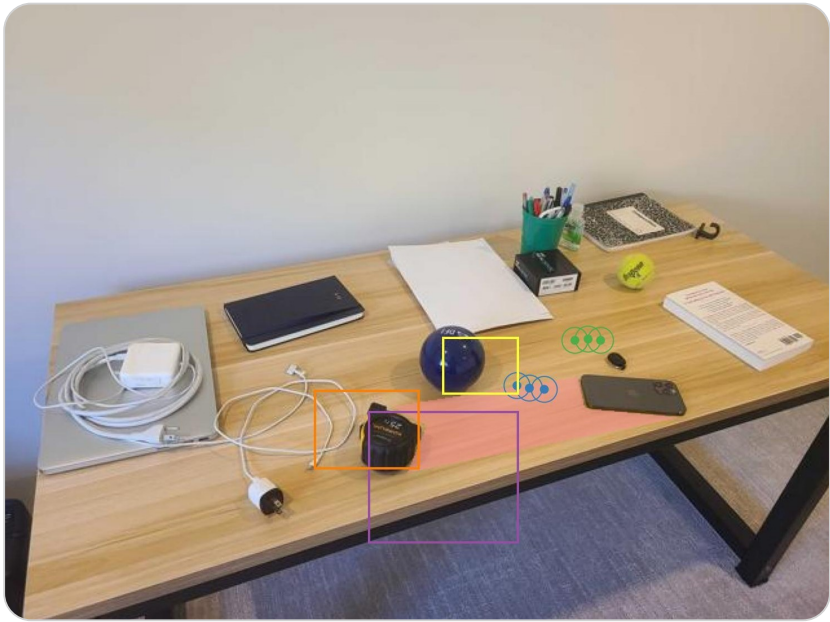

    \centering
    \includegraphics[width=1\textwidth]{Figures/supplementary/appendix_w2p_01_trace_v3.pdf}
    
    \includegraphics[width=1\textwidth]{Figures/supplementary/appendix_w2p_02_trace_v3.pdf}

    \caption{
    Qualitative results on the \textbf{Where2Place} benchmark. These examples showcase tasks that require identifying vacant space based on complex relational instructions. In both cases, TRACE's explicit Chain of Reasoning helps it correctly interpret the spatial relationship and generate precise affordance points. In contrast, baseline models either misinterpret the relation or provide less accurate localization.
}
    \label{fig:W2P}
    
\end{figure*}

To further illustrate the performance of our model, Figures \ref{fig:W2P} through \ref{fig:roboRefit} provide a qualitative comparison of \textbf{TRACE} against the \textbf{RoboPoint (Baseline)}, \textbf{SpaceLLaVA}, \textbf{GPT-4o}, and \textbf{Gemini} models. The examples are drawn from both the \textbf{Where2Place} and \textbf{RoboRefIt} benchmarks. Across these challenging scenarios, TRACE not only provides more accurate affordance points but also generates an explicit Chain of Reasoning (CoR) that makes its decision-making process transparent and interpretable. This contrasts with the baseline models, which can misinterpret spatial relations or fail to precisely locate the target region.

\begin{figure*}[ht]
    \centering
    \includegraphics[width=1\textwidth]{Figures/supplementary/appendix_Refit_01_trace_v3.pdf}
    
    \includegraphics[width=1\textwidth]{Figures/supplementary/appendix_Refit_02_trace_v3.pdf}
    
    \caption{
    Qualitative results on the \textbf{RoboRefIt} benchmark, which tests the ability to ground language instructions to specific objects in cluttered environments. TRACE successfully uses its reasoning process to identify the target objects and predict accurate affordance points. The comparison highlights how the textual rationale aids in disambiguating objects and improves the precision of the final output compared to baseline methods.
}
    \label{fig:roboRefit}
    
\end{figure*}

\section{Details on Reasoning-Augmented Data Generation}
\label{sec:dataset_details_Reasoning}

As described in the main paper, our 200,000-sample TRACE dataset is a composite. This section provides details on the generation of the novel \textbf{100,000 reasoning-augmented samples}, which form our core contribution.

These samples were created by taking image-instruction pairs from the RoboRefIt and Where2Place benchmarks (50,000 from each) and using a large language model to generate a detailed Chain of Reasoning (CoR) for each one.

The reasoning chains were generated using the \textbf{Gemini 2.5 Flash} model (`gemini-2.5-flash-lite-preview-06-17`) via its API. We prompted the model to produce a step-by-step rationale that logically connects the visual context and the language instruction to the final selection of ground-truth points. The prompt was structured to deconstruct the task into distinct cognitive steps, including: 1) Identifying reference objects in the scene, 2) Determining the goal's subtype (e.g., Placement Affordance), 3) Defining the specific target area, and 4) Explaining how the final points were generated within that area.

This process, which took approximately 40 hours of API computation time, resulted in the rich, interpretable data format shown in \Cref{fig:placement_affordance} of the main paper. This 100k subset, when combined with 100k standard instruction-following samples from VQA and LVIS, comprises the full training dataset used to fine-tune TRACE.

\section{TRACE Dataset Generation Details}
\label{sec:dataset_details}

The TRACE dataset was created to explicitly teach Vision-Language Models to generate a textual Chain of Reasoning (CoR) before predicting spatial affordances. The generation process involved leveraging a powerful large language model to create detailed rationales for existing image-instruction pairs.

First, we randomly sampled a total of 100,000 examples, drawing 50,000 from the RoboRefIt benchmark and 50,000 from the Where2Place benchmark. For each sampled image, instruction, and ground-truth affordance point set, we programmatically generated a corresponding reasoning chain.

The reasoning chains were generated using the \textbf{gemini-2.5-flash-lite-preview-06-17} model via its API. We prompted the model to produce a step-by-step rationale that logically connects the visual context and the language instruction to the final selection of points. The prompt was structured to deconstruct the task into distinct cognitive steps, including: 1) Identifying reference objects in the scene, 2) Determining the goal's subtype (e.g., Placement Affordance), 3) Defining the specific target area, and 4) Explaining how the final points were generated within that area.

This process, which took approximately 40 hours of API computation time, resulted in the rich, interpretable data format shown in \Cref{fig:placement_affordance} of the main paper. This dataset of 100,000 reasoning-augmented samples forms the core of the training data used to fine-tune TRACE.

\section{TRACE Dataset Examples}
\label{sec:dataset_examples}

\begin{figure*}[ht]
    \centering
    \includegraphics[width=1\textwidth]{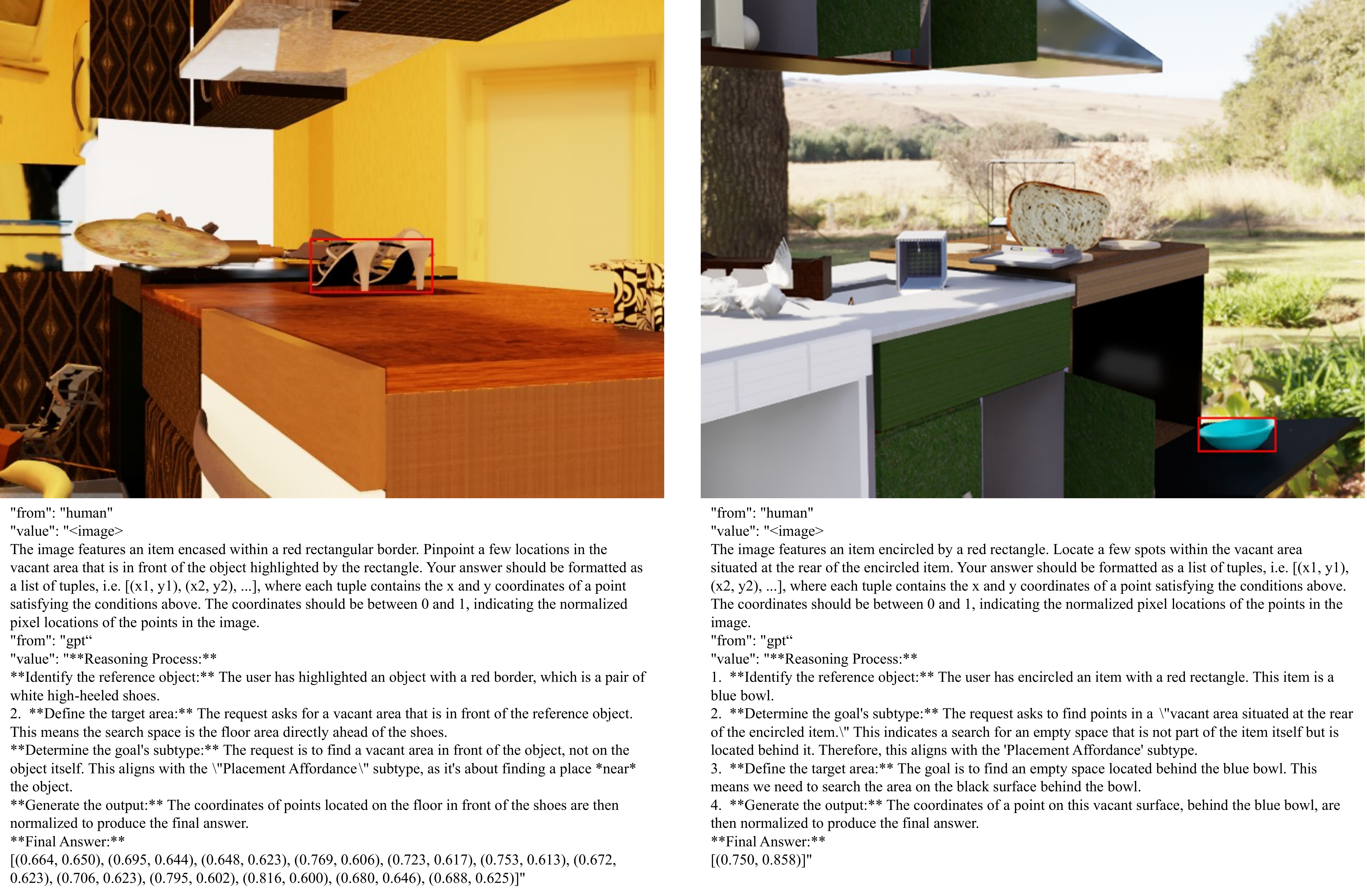}
    \caption{This image contains two data examples from the \textbf{Free Space Reference} dataset, with each example showing the prompt, the model's reasoning, and the final coordinate output.}
    \label{fig:free_space_ref}
\end{figure*}

\begin{figure*}[ht]
    \centering
    \includegraphics[width=1\textwidth]{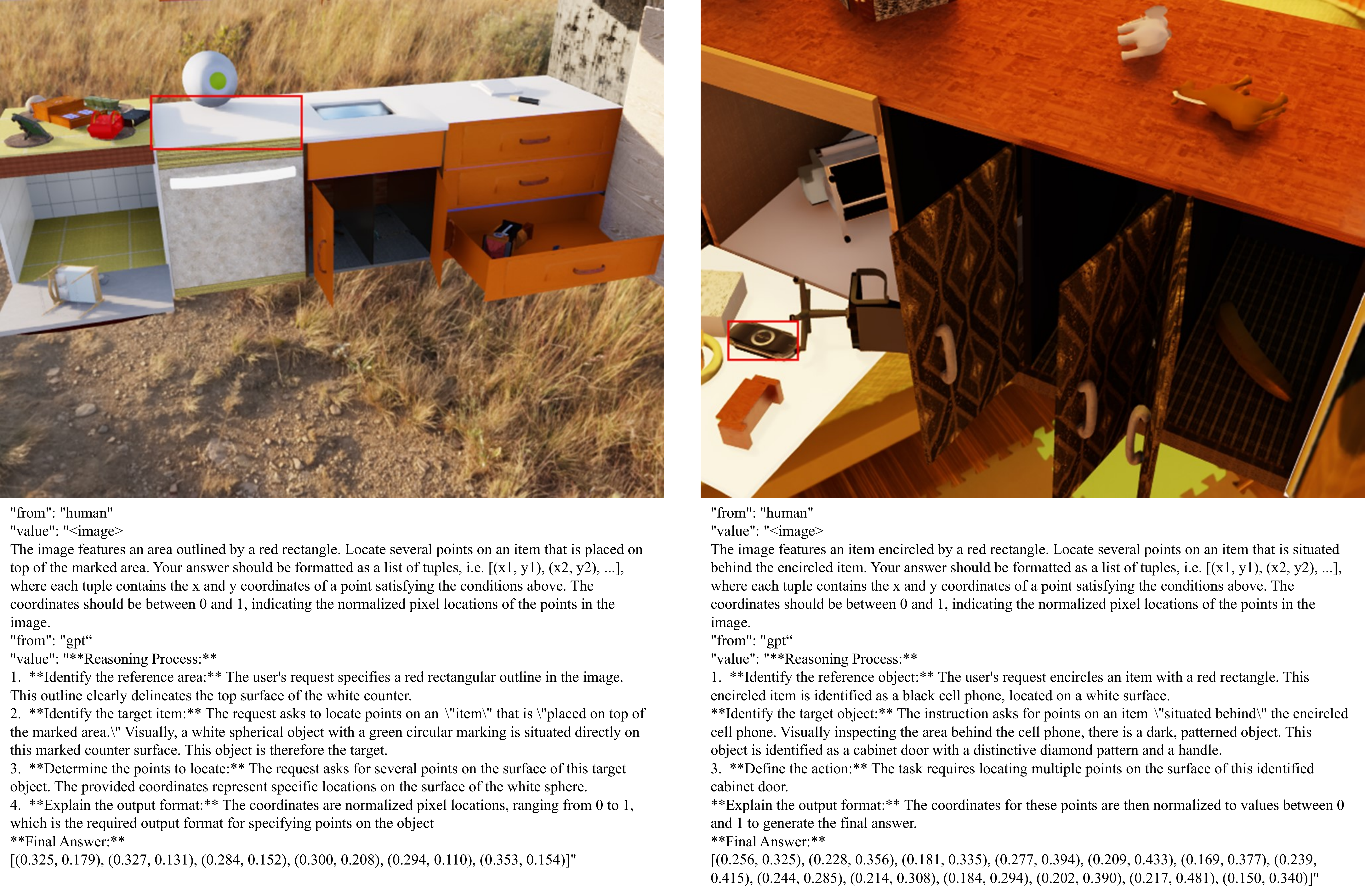}
    \caption{This image contains two data examples from the \textbf{Object Reference} dataset, with each example showing the prompt, the model's reasoning, and the final coordinate output.}
    \label{fig:object_ref}
\end{figure*}

To provide a clearer view of our training data, Figures \ref{fig:free_space_ref} and \ref{fig:object_ref} show sample data points from the TRACE dataset. Each example includes the input instruction, the synthetically generated Chain of Reasoning, and the ground-truth coordinates used for fine-tuning our model.

\end{document}